\documentclass[twocolumn]{article}

\usepackage{silence}
\usepackage{arxiv}

\usepackage[utf8]{inputenc} 
\usepackage[T1]{fontenc}    
\usepackage{hyperref}       
\usepackage{url}            
\usepackage{booktabs}       
\usepackage{amsfonts}       
\usepackage{nicefrac}       
\usepackage{microtype}      
\usepackage{lipsum}		
\usepackage{graphicx}
\usepackage[numbers]{natbib}
\usepackage{doi}
\usepackage{multicol}
\usepackage{amsmath}

\usepackage{physics}
\usepackage{geometry}

\title{Denoising Diffusion Probabilistic Models as a Defense against Adversarial Attacks}

\author{ 
    Lars Ankile\\
Harvard University\\
\texttt{larsankile@g.harvard.edu} \\
\And
    Anna Midgley \\
Harvard University\\
\texttt{amidgley@g.harvard.edu}
\And
Sebastian Weisshaar \\
Harvard University \\
\texttt{sweisshaar@g.harvard.edu} \\
}

\renewcommand{\headeright}{}
\renewcommand{\undertitle}{}
\renewcommand{\shorttitle}{Denoising Diffusion Probabilistic Models as a Defense against Adversarial Attacks}

\hypersetup{
pdftitle={Denoising Diffusion Probabilistic Models as a Defense against Adversarial Attacks},
pdfsubject={cs.ML},
pdfauthor={Lars Ankile, Anna Midgley, Sebastian Weisshaar},
pdfkeywords={diffusion, robust ml, adversarial training},
}

\begin{document}

\twocolumn[

\maketitle

\begin{abstract}
    \noindent
    Neural Networks are infamously sensitive to small perturbations in their inputs, making them vulnerable to adversarial attacks. This project evaluates the performance of Denoising Diffusion Probabilistic Models (DDPM) as a purification technique to defend against adversarial attacks. This works by adding noise to an adversarial example before removing it through the reverse process of the diffusion model. We evaluate the approach on the PatchCamelyon data set for histopathologic scans of lymph node sections and find an improvement of the robust accuracy by up to 88\% of the original model's accuracy, constituting a considerable improvement over the vanilla model and our baselines. The project code is located at \url{https://github.com/ankile/Adversarial-Diffusion}.
\end{abstract}
\vspace*{6mm}
]

\raggedbottom

\section{Introduction \& Related Work}
Imperceptible perturbations to the input $X$ to a Neural Network (NN) can deceive the most accurate models into predicting the incorrect class with high confidence \cite{szegedy2013intriguing}. Two main strategies exist to defend against such adversarial attacks \cite{DiffPure}. The first is adversarial training which trains NNs on adversarial examples (i.e., samples specifically constructed to deceive a given classifier). However, this method can only defend against the attack types one trained the model to withstand. The alternative technique uses Generative Models (GMs) to purify images by removing the adversarial perturbation before passing the image to the classifier. This method is a more general defense against adversarial attacks and can handle unseen threats. However, due to the shortcomings of GMs, the technique currently performs worse than adversarial training methods. Shortcomings depend on the type of GM used but include mode collapse, low sample quality, and lack of proper randomness \cite{DiffPure}.

Recently, diffusion models have emerged as one of the most powerful GMs, capable of overcoming the mentioned shortcomings \cite{DiffusionPaper}. In this project, we propose that diffusion models can be used for adversarial purification and are a natural fit for this purpose. The rationale is that the diffusion model gradually adds noise to the input in the forward process, perturbing the data. The \textit{a priori} reason this would defend against adversarial attacks is that the adversarial perturbations are disrupted by the added noise and hence cannot disturb the classifier. 

The forward noise process in the DDPM is described by,

\begin{align}
    q_t\left(x_t \mid x_0\right)=N\left(x_t \mid \sqrt{\bar{\alpha}_t} x_0,\left(1-\bar{\alpha}_t\right) I\right),
\end{align}

where $\bar{\alpha}_t=\prod_{s=1}^t\left(1-\beta_t\right)$ and $\beta_t$ defines the `noise schedule,' i.e., how much noise one adds at each step. Then, in the reverse process, the model aims to take the noisy image and remove the noise to retrieve the original input, thereby learning to recover the input. The reverse process is the joint distribution, $p_\theta\left(\mathbf{x}_{0: T}\right)$, which is as a Markov chain with learned transitions starting at $p(x_T)$,

\begin{align}
    p_\theta\left(x_{0: T}\right)=p\left(x_T\right) \prod_{t=1}^T N\left(x_{t-1} \mid \mu_\theta\left(x_t, t\right), \beta_t I\right).
\end{align}

The mean $\mu_\theta\left(x_t, t\right)$ is an NN parameterized by $\theta$ \cite{DiffusionPaper}. One finds the generative model by maximizing the Evidence Lower Bound Criterion (ELBO) \cite{DiffusionPaper}. The reverse process is intuitively similar to purification, where the goal is to remove a perturbation from an adversarial example. Researchers have empirically demonstrated that diffusion models can generate high-quality samples \cite{DiffusionPaper}. This ability ensures that the cleaned image follows the original distribution of the data. Furthermore, reliable attacks against diffusion models are harder to design as the model is stochastic. These properties are beneficial for adversarial purification.

\emph{DiffPure} uses diffusion models for adversarial purification \cite{DiffPure}. The method first adds a small amount of noise to an adversarial example with the forward diffusion process and then recovers the purified, clean image through the reverse process. The hope is that the perturbations gradually mix with noise and that the added Gaussian noise dominates them. The reverse process then removes both the added noise and the perturbations. In \cite{DiffPure}, the researchers find the method to outperform current adversarial training and adversarial purification methods on three image data sets---CIFAR-10, ImageNet, and CelebA-HQ---with three classifier architectures---ResNet, WideResNet, and ViT.

\textbf{Contributions:} Our project complements the results in \cite{DiffPure} by applying a purifying diffusion model to the classification of metastatic tissue. In the biomedical sciences, model robustness is paramount, and research into making current methods more reliable in this setting is essential. Furthermore, as \cite{DiffPure} points out, one weakness of the diffusion method is that it requires many inference steps and is slow. Our proposed solution shows strong results using a low noise level of $t^*=0.04$ (40 inference steps), which is smaller than in previous works, leading to faster inference.

\section{Data, Models, \& Methods}
\label{sec:methods}

\textbf{Data sets and network architectures:} We used the PatchCamelyon data set to evaluate our proposed method. The data set contains histopathologic scans of lymph node sections. Each image has a binary label that indicates the presence of metastatic tissue, i.e., cancerous cells \cite{Pocock2022}. The binary classes are perfectly balanced. For the classifier, we consider the widely used \texttt{ResNet} model and specifically use \texttt{ResNet101}, which we refer to as \texttt{ResNet} \cite{ResNet}. In addition, we used TiaToolBox's \texttt{ResNet} model pre-trained on the PCam data. For the robust adversarial training, we experimented with many architectures and found \texttt{GoogLeNet}\cite{szegedy2015going} to be the only one that did not collapse to the naïve solution.

\textbf{Adversarial attacks:}
We chose to use adaptive attacks designed with full knowledge of the model's defense. The adversarial examples were created by finding the perturbation from the set of allowable perturbations that maximized the loss given by the following equation,

\begin{align}
    \underset{\delta \in \Delta}{\operatorname{maximize}} \: \ell\left(h_\theta(x+\delta), y\right).
\end{align}

We found the maxima with project gradient descent on the common perturbation set of $\ell_{\infty}$ ball, defined as $\Delta=\left\{\delta:\|\delta\|_{\infty} \leq \epsilon\right\}$. The restriction of the norm of the perturbation ensures that the perturbed image is indistinguishable from the original image. Thus, it would still be reasonable to expect the correct classification. We reasoned that the value of $\epsilon$ should be greater than pixel 1 unit of change from the original image to ensure that it is not possible to recover the original image by rounding the pixel values in the perturbed image. $\epsilon$ was set to be $2/255 = 0.0078$, which was held constant throughout the experiments. At this level, the adversarial examples consistently fooled the classifier while maintaining the imperceptibility of perturbation.

\textbf{Proposed method:}
The proposed method is a diffusion model coupled with a classifier. We used the \texttt{diffusers} library made by Hugging Face \cite{von-platen-etal-2022-diffusers}. The type of diffusion model is a DDPM. The scoring model was a \texttt{U-Net} trained from scratch with a linear noise schedule on \textasciitilde 30k tissue sample images for 15 epochs, taking a couple of hours on a single Colab GPU. We note that there is ample room for improvement in the diffusion purification model regarding the number of epochs, the number of samples, data augmentation, and hyperparameter tuning. However, we could not optimize these values more during the project due to a lack of computing resources.

\textbf{Baseline models:} We compared our proposed method of adversarial defense to two other defense methods, which provide performance baselines. As a first baseline, we used a simple Gaussian noise procedure followed by the classifier, referred to as \texttt{NoiseDefense}. The noise added is equivalent to the forward process of the diffusion model. In theory, these noisy images should contain enough information for correct classification while having enough noise to distort the precisely constructed adversarial perturbations. If the diffusion model can extract enough information to correctly denoise the image, the classifier could also use this information.

\begin{figure}
    \includegraphics[width=0.45\textwidth]{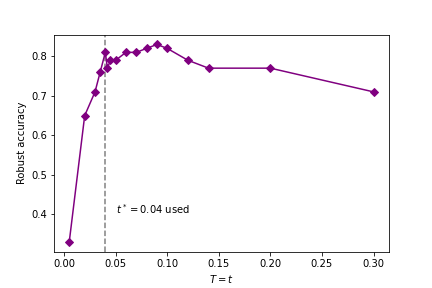}
    \caption{Robust accuracy for different noise levels $t\in[0.001, 0.300]$ for a subset of the validation data.}
    \label{fig:noise-level}
\end{figure}

We decided what noise level to add to the adversarial images using cross-validation on a subset of the data. We used different $t$ for the forward process $q_t(x_t|x_0)$ and evaluated the classifier's performance on the DDPM de-noised images. We found an optimal accuracy-speed performance at $t^*=0.04$. See the next section and \autoref{fig:noise-level}.

As the second baseline, we trained a robust classifier using adversarial examples created during every weight update of the model. Thus, we can interpret each model update as a min-max problem where we first maximize the loss w.r.t input perturbations $x+\delta$ and then minimize the loss w.r.t the model weights.

\textbf{Optimal noise level:}
The noise level is an important metric in determining the performance of the diffusion model in adversarial purification. \autoref{fig:noise-level} shows the accuracy of ResNet101 after noising and denoising adversarial examples with different noise levels $t\in[0, 1]$. There are several noteworthy results in this graph. First, a wide range of time steps results in similar accuracy. Furthermore, the level of noise that can be added to an image before the performance drops is quite significant. An image at $t=0.10$ looks very noisy to the human eye, yet the diffusion process can recover relevant features such that the classifier can still detect metastatic tissue. With increasing noise, though, there is a drop in performance when exceeding $t=0.20$ and beyond. Lastly, this graph shows that we must add a significant amount of noise to counter the adversarial perturbations effectively. Despite good robust accuracy results for $t=0.10$, we can speed up inference with 60\% by choosing a noise level at the lower end of the optimal range. We, therefore, chose $t^*=0.04$.

\section{Experimental Results}

\begin{figure}
    \includegraphics[width=0.45\textwidth]{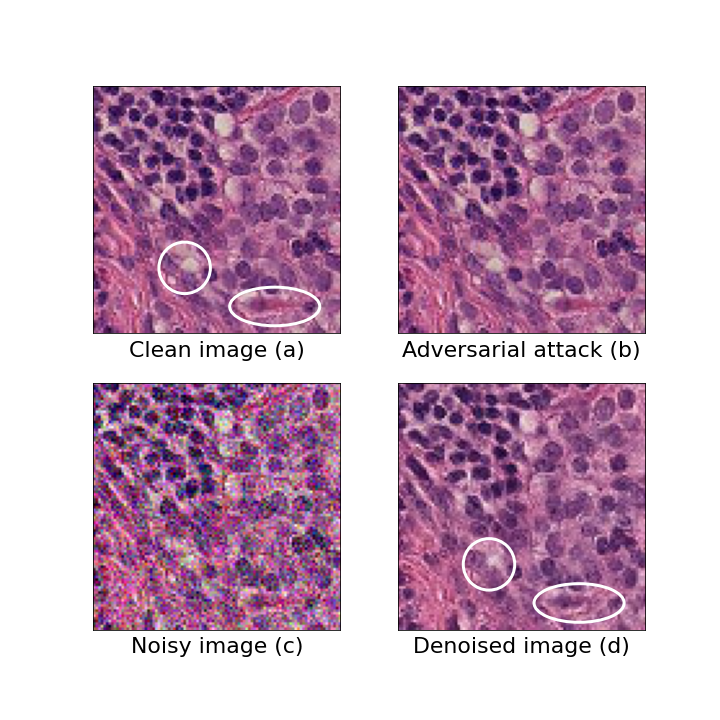}
    \caption{An example of a tissue sample in the different stages of the model pipeline.}
    \label{fig:tissue-img}
\end{figure}

\textbf{Visualization of pipeline model outputs:}
\autoref{fig:tissue-img} shows an example of the outputs of our pipeline. Image (a) shows the original histopathologic scan of a lymph node with metastatic tissue (correctly classified by classifier). Image (b) shows an adversarial example. Even though there is no discernible difference for the human eye, the \texttt{ResNet} model classifies the upper left image with probability $P(Y=0|X)=0.9997$ of no metastatic tissue present. Image (c) shows the noise added in the \texttt{NoiseDefense} and as preparation for the reverse diffusion process. Image (d) shows the tissue after the diffusion model removed the noise. Note that this process destroyed some details in the image. The white circles show differences between the original image (a) and the purified image (d), which ideally should be identical.

\textbf{Adversarial perturbations:}
In creating the adversarial examples, we found that all perturbations were on the $L_\infty$ norm ball perimeter, as theory would predict. Using an $L_\infty$ norm ball as a restriction causes all the perturbations to take the maximum allowed step in the direction of the gradient. As more significant changes are most likely to damage the classifier, all perturbations lie on the constraint boundary, equal to the maximum allowable norm of perturbation.

\begin{figure}
    \includegraphics[width=0.45\textwidth]{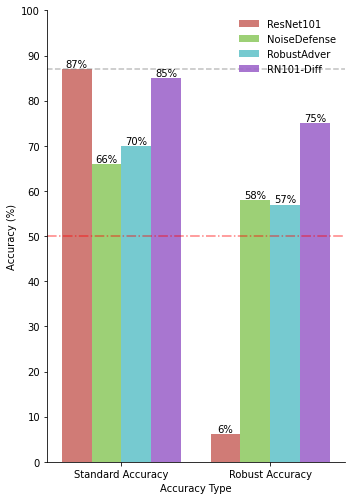}
    \caption{The results of running our four models on 1000 test samples for both standard accuracy (left) and robust accuracy (right). The vanilla \texttt{ResNet} model is red, and our method is purple. It is also important to note that the robust adversarially trained model is an instance of a \texttt{GoogLeNet}, and not \texttt{ResNet}, as this was the only tested architecture that generalized under adversarial training.}
    \vspace*{-1mm}
    \label{fig:results}
\end{figure}

We present the results from our experiments in \autoref{fig:results} and will analyze the different models' performance on the test data in the following paragraphs.

\textbf{Vanilla classifier:}
To begin with, the pre-trained \texttt{ResNet} model performed at an 87\% accuracy on the standard test data. This result will serve as the north star accuracy any adversarially robust method would want to achieve in our context. Next, we observe that adversarial attacks at a $\norm{\epsilon}_\infty\leq\frac{2}{255}$ level are highly effective and result in an adversarial accuracy of 6\%, which is worse than the 50\% the naïve baseline would achieve (since the data set is balanced). This result shows that creating adversarial examples for this data set works with near certainty.

\textbf{Noise Defense:}
The first baseline model, the \texttt{NoiseDefense}, tries to use noise to `wash out' the targeted adversarial attacks. The hope is that there is a level of noise that undermines the targeted attack without undermining the classifier. From the green bars in \autoref{fig:results}, we see that the noise did undermine the adversarialness of the attacks by observing that the robust accuracy increased from 6\% to 58\% with the added noise. However, we also observe that the standard accuracy falls to 66\%, i.e., marginally better than the naïve. This result indicates that the added noise removes much of the signal. However, the perturbations were not sufficiently removed at a lower level of added noise to allow the model to predict better than chance.

\textbf{Adversarially trained classifier:}
The cyan bar in \autoref{fig:results} shows the performance of the adversarially trained robust model. This model performs marginally better than the \texttt{NoiseDefense} model, with a 70\% standard accuracy and 57\% robust accuracy. In the process of training this model, we observed several notable findings. First, there seems to be an inherent trade-off between standard and robust accuracy, as whenever we observed an increase in robust accuracy, it came with a drop-off in standard accuracy, which is also supported by the literature \cite{raghunathan2020understanding}. This trade-off could also be made sense of in the context of the increased sample complexity that adversarially robust models have \cite{schmidt2018adversarially}. The robust model needs to balance performance on standard and adversarial images while dealing with the sample complexity increased due to the addition of adversarial examples.

Third, robust adversarial training is computationally expensive and finicky. Since one is solving one optimization problem for each training step, the process is costly. Furthermore, the models we trained were sensitive to hyperparameter choice and tended to collapse toward the naïve solution (predicting a class at random) if trained for too long. Despite extensive hyperparameter search, the standard accuracy dropped off markedly in our experiments. Out of the seven different architectures we tested, only \texttt{GoogLeNet} did not collapse to the naïve solution. Lastly, adversarially robust training is model-specific and assumes a particular attack type, i.e., how one trains is how the model will be made more robust.

\textbf{Diffusion model coupled with classifier:}
\autoref{fig:results} shows the results of our diffusion approach in purple. The first thing to note is how close the standard accuracy is to the vanilla model, indicating that the diffusion process is successfully recreating images that are faithful to the original (i.e., not losing the essential details that allow the classifier to discriminate the classes). Accurate reconstruction is an important attribute, as a robust model with low accuracy on non-perturbed data cannot be used in practice. Especially for a life-critical application like the detection of metastatic tissue, high performance on non-adversarial examples is crucial.

Furthermore, the adversarial accuracy is also comparably high at 75\% accuracy, vastly better than the baselines and the vanilla model. Again, the diffusion model can accurately purify the images (both added noise and adversarial perturbations) while retaining the crucial details. This ability could be better, though, as we did not achieve equal robust accuracy to the standard accuracy. As discussed in \autoref{sec:methods}, the chosen noise level $t^*=0.04$ was used as it provided fast inference and high precision.

In contrast to robust adversarial training, the diffusion defense does not assume any specific attack and the diffusion model pairs with any model without any extra fitting. Though we did not test it in this work, there is reason to believe that the diffusion defense would be effective against perturbations within a differently-sized norm ball, as the diffusion defense is a standard diffusion model.

\textbf{Difficulties in detecting metastatic tissue:}
The PCam data set classifies images as class \texttt{1} if at least one pixel with cancer is present in the center $32\times 32$ pixel region of the picture. This property means that the data set is sensitive to small changes. The difficulty in predicting such small areas of cancer could explain the poor performance of the adversarially robust trained model. The model must be susceptible to minor differences in the images as these small changes indicate metastatic tissue. If we introduce perturbations, the model cannot disregard them, leading to poor performance. If we compare this with a more general classification task like CIFAR-10, we see that this issue does not exist \cite{DiffPure}. In CIFAR-10, a small perturbation is generally not decisive in distinguishing a cat from an airplane. Therefore robust training can find workarounds for adversarial examples. In our data set, this luxury does not exist, as the sole purpose of this classification task is to find small perturbations.

\section{Conclusion \& Future Work}
This work showed that diffusion models are effective in adversarial purification. Furthermore, their coupling with a classifier increases the overall robustness of the pipeline to adversarial attacks. Thus potentially setting a new gold standard in adversarial defense. In addition, we observed better results for standard accuracy and robust accuracy than the baseline approaches.

There are several avenues for future research on this topic. For one, we want to explore whether it is possible to steer the diffusion model during training toward outputting a cleaned image that is easily classified. One would not train the classifier and diffusion model independently but instead train in conjunction, utilizing shared information to improve the defense. However, by including the classifier in the training of the diffusion model, the defense would no longer be model agnostic.

Secondly, while training the diffusion model, we realized that when the noise exceeded a threshold, the diffusion model outputted an image quite different from the input. We hypothesize that when the noise is too large, the signal is low, and the model cannot correctly rebuild the aspects of the image relevant for classification. Future research could analyze how more training on the diffusion process would enable the defense to diffuse even more noise from the image.

\onecolumn

\bibliographystyle{abbrv}
\bibliography{references}  

\end{document}